\pdfoutput=1

\documentclass[11pt]{article}

\usepackage[preprint]{acl}

\usepackage{times}
\usepackage{latexsym}

\usepackage[T1]{fontenc}

\usepackage[utf8]{inputenc}

\usepackage{microtype}

\usepackage{inconsolata}

\usepackage{graphicx}

\usepackage{microtype}
\usepackage{graphicx}
\usepackage{array}
\usepackage{booktabs}
\usepackage{multirow}
\usepackage{siunitx}
\usepackage{booktabs} 
\usepackage{graphicx}
\usepackage{subcaption}
\usepackage{mwe}
\usepackage{graphicx}

\usepackage{xcolor}
\usepackage{colortbl}
\usepackage[linesnumbered,ruled,vlined]{algorithm2e}

\usepackage{amsmath}
\usepackage{amssymb}

\definecolor{goodgreen}{RGB}{34, 139, 34}
\definecolor{badred}{RGB}{220, 20, 60}

\usepackage{dblfloatfix}

\usepackage{amsmath}
\usepackage{amssymb}
\usepackage{mathtools}
\usepackage{amsthm}
\usepackage{enumitem}
\usepackage{svg}
\usepackage{tcolorbox}
\usepackage{CJKutf8}
\usepackage{authblk}
\setlist[itemize]{itemsep=-1mm, topsep=-1mm}

\newtcolorbox{supbox}[2][]{colback=black!2!white,colframe=cyan!80!black,
title={#2},#1}
\newtcolorbox{membox}[2][]{colback=black!2!white,colframe=orange!90!white,
title={#2},#1}
\newtcolorbox{egbox}[2][]{ colback=black!2!white,colframe=green!60!black,
title={#2},#1}
\newtcolorbox{prtbox}[2][]{ colback=black!2!white,colframe=violet!80!white,
title={#2},#1}

\title{Talk Structurally, Act Hierarchically: \\A Collaborative Framework for LLM Multi-Agent Systems}

\author{
    Zhao Wang$^{*,\dag}$, 
    Sota Moriyama$^*$, 
    Wei-Yao Wang, 
    Briti Gangopadhyay, 
    Shingo Takamatsu
}

\affil{Sony Group Corporation, Japan}

\begin{document}
\begin{CJK}{UTF8}{min}  
\maketitle

\def\thefootnote{*}\footnotetext{These authors contributed equally to this work} 
\def\thefootnote{\dag}\footnotetext{Corresponding author: Zhao Wang (Email Address: Zhao.Wang@sony.com)} 

\begin{abstract}
Recent advancements in LLM-based multi-agent (LLM-MA) systems have shown promise, yet significant challenges remain in managing communication and refinement when agents collaborate on complex tasks. In this paper, we propose \textit{Talk Structurally, Act Hierarchically (TalkHier)}, a novel framework that introduces a structured communication protocol for context-rich exchanges and a hierarchical refinement system to address issues such as incorrect outputs, falsehoods, and biases. \textit{TalkHier} surpasses various types of SoTA, including inference scaling model (OpenAI-o1), open-source multi-agent models (e.g., AgentVerse), and majority voting strategies on current LLM and single-agent baselines (e.g., ReAct, GPT4o), across diverse tasks, including open-domain question answering, domain-specific selective questioning, and practical advertisement text generation. These results highlight its potential to set a new standard for LLM-MA systems, paving the way for more effective, adaptable, and collaborative multi-agent frameworks. 
The code is available at \href{https://github.com/sony/talkhier}{https://github.com/sony/talkhier}.
\end{abstract}

\section{Introduction}
\label{Intro}
Large Language Model (LLM) Agents have broad applications across domains such as robotics~\citep{brohan2022code}, finance~\citep{shah2023finGPT,zhang2024finagent}, and coding~\cite{chen2021evaluating,hong2023metagpt}. By enhancing capabilities such as autonomous reasoning~\citep{wang2024rethinking} and decision-making~\citep{eigner2024determinants}, LLM agents bridge the gap between human intent and machine execution, generating contextually relevant responses~\citep{arXiv2024_Survey-MultiAgent_2}.
\begin{figure}[t] 
\centering
\includegraphics[width=\linewidth]{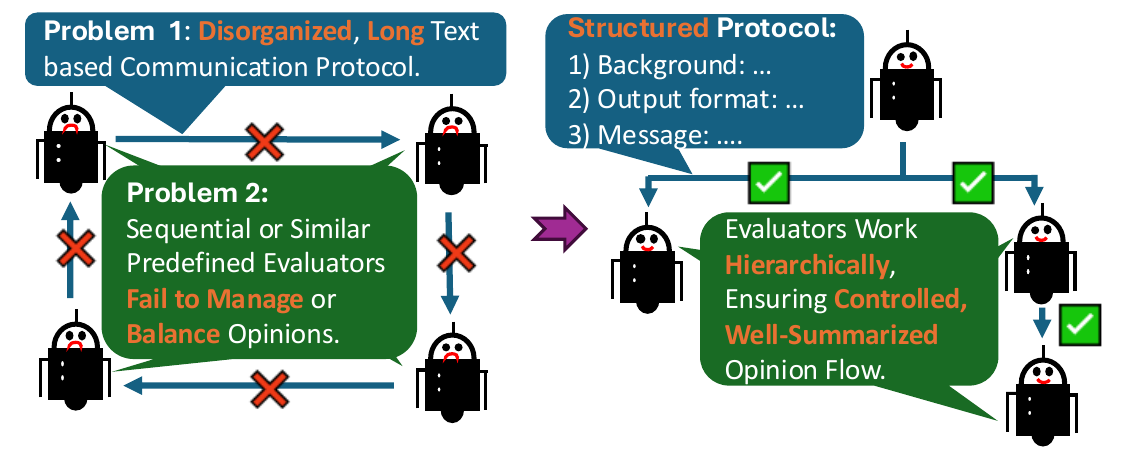}
\caption{Existing LLM-MA methods (left) face two major challenges: 1) disorganized, lengthy text-based communication protocols, and 2) sequential or overly similar flat multi-agent refinements. In contrast, \textit{TalkHier} (right) introduces a well-structured communication protocol and a hierarchical refinement approach.}
\label{fig:teaser1}
\end{figure}

\begin{figure}[t] 
\centering
\hspace{-15pt} 
\includegraphics[width=\linewidth]{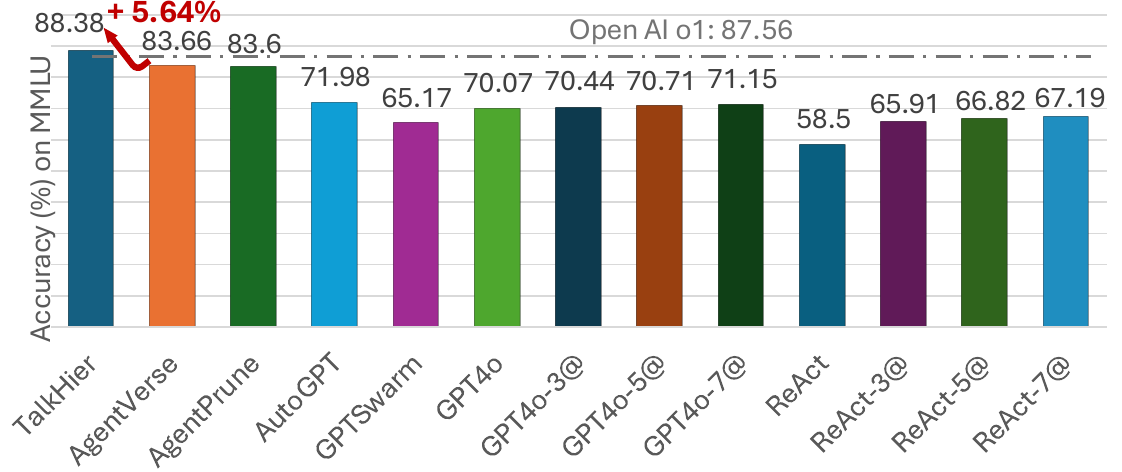}
\caption{Our \textit{TalkHier} built on GPT4o surpasses inference scaling models (OpenAI-o1), open-source multi-agent models (AgentVerse and etc.), and models with majority voting strategies (ReAct, GPT4o) on five subtasks of MMLU.}
\label{fig:teaser2}
\end{figure}
Recent research has primarily focused on LLM-based Multi-Agent (LLM-MA) systems, which leverage collective intelligence and specialize each agent with the corresponding subtasks, to solve complicated and multi-step problems.
For instance, previous works on LLM-MA have explored approaches where instances of LLMs, referred to as agents~\cite{arXiv2023_Survey-Agent_2, arXiv2023_Survey-Agent_3, FCS2024_Survey-Agent, arXiv2024_Survey-Agent_4, arXiv2024_Survey-Agents-CompExp}, collaborate synergistically by debate \cite{chen2024comm}, reflection \cite{he2024enhancing}, self-refinement \cite{madaan2023selfrefine}, or multi-agent based feedback refinement \cite{yang2023idea2img}.
These systems employ diverse communication topologies to enable efficient interactions between agents such as Chain~\cite{cot} and Tree~\cite{tot} structures, among others~\cite{qian2024scaling, zhuge2024gptswarm, zhang2024cut}.

Despite the promising advancements in LLM-MA systems, several challenges in this field remain unexplored (shown in Figure~\ref{fig:teaser1}):

\noindent\textbf{1) Disorganized communication in text form.}
Agents often engage in debates~\cite{zhao2023competeai}, share insights~\cite{chen2024comm}, or perform refinement~\cite{madaan2023selfrefine, yang2023idea2img} to effectively solve complex tasks, with their exchanges primarily in text form~\cite{guo2024large}.
However, communication often becomes disorganized because it requires explicitly describing agent tasks, providing background context for the communication, and specifying the required output formats. These factors together lead to lengthy and unstructured exchanges, making it difficult for agents to manage subgoals, maintain output structures, and retrieve independent memories from prior actions and observations.

\noindent\textbf{2) Refinement schemes.}
While some studies have shown that incorporating agent debates~\cite{chen2024comm} or evaluation-based multi-agent refinement~\cite{wang2023coeval, yang2023idea2img} can improve system accuracy, these approaches also expose significant limitations. As the number of agents increases, LLM-MA systems face challenges in effectively summarizing opinions or feedback~\cite{fang2024multi}. They often fail to balance these inputs, frequently overlooking some or exhibiting biases based on the order in which feedback is provided~\cite{errica2024did}.

In this paper, we propose a novel collaborative LLM-MA framework called \textit{Talk Structurally, Act Hierarchically (TalkHier)}-the first collaborative LLM-MA framework to integrate a well-structured communication protocol with hierarchical refinement.
Our key contributions shown in Figure~\ref{fig:teaser1} and \ref{fig:teaser2} are as follows:

\begin{enumerate}[leftmargin=*, itemsep=0pt, topsep=0pt]
    \item \textbf{Well-Structured, Context-Rich Communication Protocol:} \textit{TalkHier} introduces a novel communication protocol that incorporates newly proposed elements: \textit{messages}, \textit{intermediate outputs}, and relevant \textit{background information}. These components form the foundation of a well-structured protocol that organizes agent communication, ensuring clarity and precision. By embedding these elements, \textit{TalkHier} significantly improves communication accuracy and efficiency compared to traditional text-based methods.

    \item \textbf{Hierarchical Refinement in LLM-MA Systems:} \textit{TalkHier} enhances traditional multi-agent evaluation systems with a hierarchical refinement framework, 
    enabling agents to act hierarchically. This approach addresses such as the difficulty in summarizing opinions or feedback as the number of agents increases, balancing diverse inputs, and mitigating biases caused by the order of feedback processing, resulting in more reliable and robust interactions.

    \item \textbf{State-of-the-Art Results Across Benchmarks:} Experimental results show that \textit{TalkHier} achieves state-of-the-art performance on diverse benchmarks, including selective problem-solving in complex sub-domains, open question answering, and Japanese text generation tasks. Ablation studies confirm the effectiveness of each component, demonstrating their contributions to the framework’s overall success.
\end{enumerate}

\begin{figure*}[!h]
\centering
\includegraphics[width=0.93\linewidth]{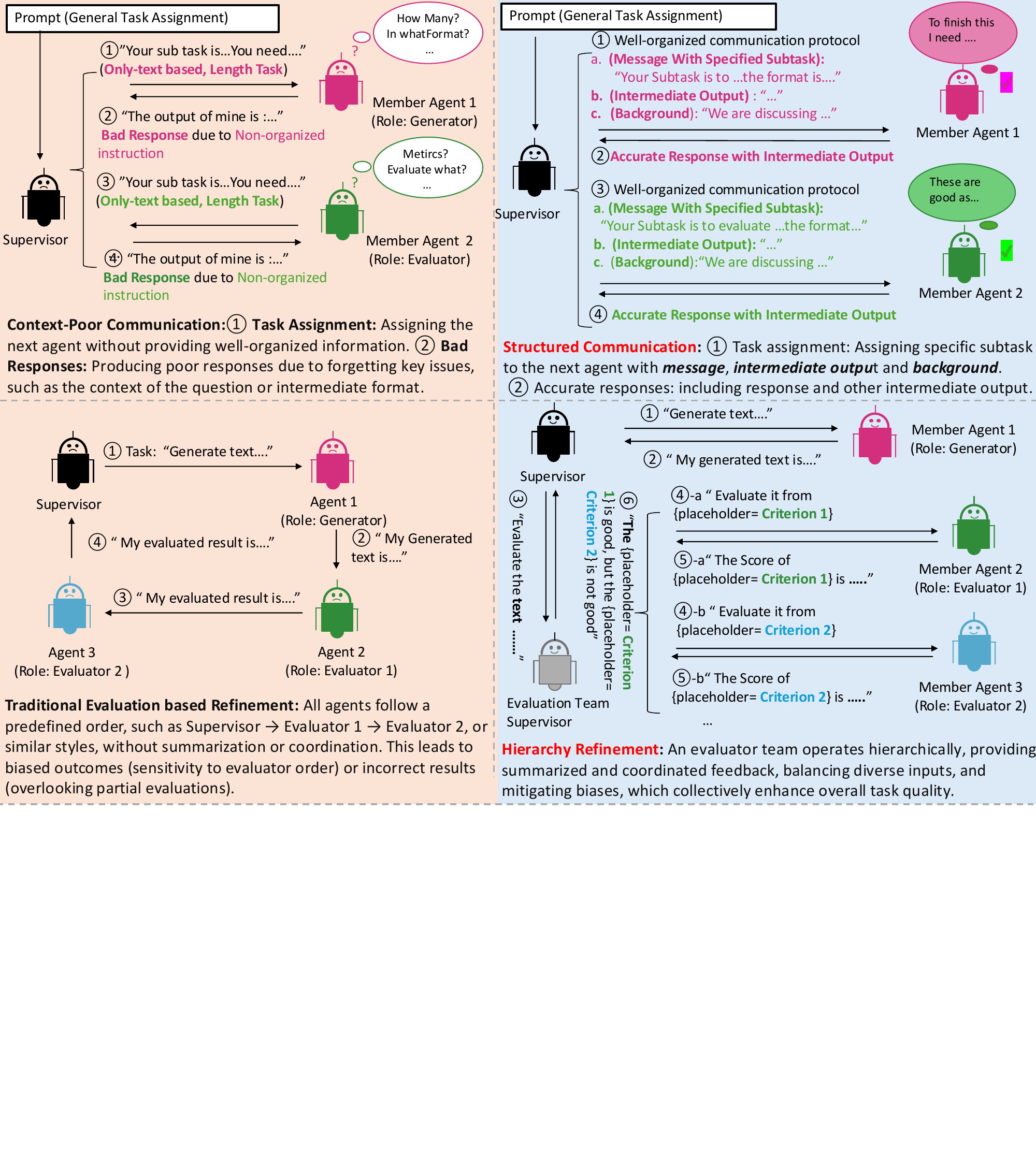}
\vskip -150pt
\caption{Comparisons between existing approaches (left) and ours (right). Our \textit{TalkHier} proposes a new communication protocol (first row) featuring context-rich and well-structured communication information, along with a collaborative hierarchical refinement (second row) where evaluations provide summarized and coordinated feedback within an LLM-MA framework.}
\label{fig:architecture}
\end{figure*}

\section{Related Work}
\label{rw}

\paragraph{Collaborative LLM-MA.}
LLM-MA systems enable agents to collaborate on complex tasks through dynamic role allocation, communication, and task execution~\cite{guo2024large, han2024challenges}. Recent advancements include agent profiling~\cite{yang2024multi}, hierarchical communication~\cite{rasal2024harmony}, and integration of reasoning and intentions~\cite{qiu2024collaborative}. 
However, challenges remain in ensuring robust communication, avoiding redundancy, and refining evaluation processes~\cite{talebirad2023collaboration}. Standardized benchmarks and frameworks are needed to drive future progress~\cite{li2024survey}.

\paragraph{Communication in LLM-MA.}
Effective communication is crucial for collaborative intelligence~\cite{guo2024large}. While many previous works, including chain~\citep{cot}, tree~\citep{tot}, complete graph~\citep{qian2024scaling}, random graph~\citep{qian2024scaling}, optimizable graph~\citep{zhuge2024gptswarm}, and pruned graph~\citep{zhang2024cut} methods have focused on communication topologies, there has been limited discussion on the optimal form of communication. Most systems rely on text-based exchanges~\cite{zhang2024cut, shen2024small}, which is inefficient and prone to errors as agents often lose track of subtasks or fail to recall prior outputs as tasks grow in complexity. We argue for structured communication protocols that guide subtasks with clear, context-specific instructions, ensuring coherence across interactions.

\paragraph{Feedback-Based Refinement.}
Feedback mechanisms, such as Self-Refine~\cite{madaan2023selfrefine} and generator-evaluator frameworks~\cite{wang2023coeval}, improve system accuracy through iterative refinement. However, these methods face challenges in managing diverse feedback, which can lead to bias or inefficiencies if inputs are not well-organized~\cite{xu2024cooperative}. Scalable, unbiased solutions are essential to enhance multi-agent evaluation processes.

\section{Methodology}

\textit{TalkHier} aims to design a LLM-MA system represented as a graph $\mathcal{G} = (\mathcal{V}, \mathcal{E})$, where $\mathcal{V}$ denotes the set of agents (nodes) and $\mathcal{E}$ represents the set of communication pathways (edges). Given an input problem $p$, the system dynamically defines a set of communication events \( \mathcal{C}_p \), where each event \( c_{ij}^{(t)} \in \mathcal{C}_p \) represents a communication between agents \( v_i \) and \( v_j \) along an edge \( e_{ij} \in \mathcal{E} \) at time step \( t \).
While the graph structure \( \mathcal{G} \) remains fixed, the communication events \( \mathcal{C}_p \) are dynamic and adapt to the specific task.

\begin{figure*}[t]
    \centering
    \begin{supbox}{ Supervisor Prompt Template}
    \small
    Team Members: [\small \textit{Description of each team member's role}]\\
    Conversation History: [\textit{Independent Conversation History}]\\
    Given the conversation above, output the following in this exact order:\\
    1. `thoughts': Output a detailed analysis on the most recent message. In detail, state what you think should be done next, and who you think you should contact next. \\
    2. Who should act next? Select one of: [\textit{Team member names}] and output as `next'. When you have determined that the final output is gained, report back with FINISH.\\
    3. `messages': If the next agent is one of [\textit{Team member names}], give detailed instructions. If FINISH, report a summary of all results.\\
    4. The detailed background of the problem you are trying to solve (given in the first message) as `background'.\\
    5. The intermediate outputs to give as `intermediate\_output'.
    \end{supbox}
    \begin{membox}{ Member Prompt Template}
    \small
    [\textit{Role of member}]\\
    Background: [\small \textit{Background information given by Supervisor}]\\
    Conversation History: [\textit{Independent Conversation History}]
    \end{membox}
    \caption{Prompts for acquiring the contents of the context-rich, structured communication protocol in \textit{TalkHier}.}
    \label{fig::com_prompt}
\end{figure*}

\subsection{Agents with Independent Memory}

Each agent \( v_i \in \mathcal{V} \) in graph \( \mathcal{G} \) can be formally represented as:
\begin{equation*}
v_i = \left( \texttt{Role}_i, \texttt{Plugins}_i, \texttt{Memory}_i, \texttt{Type}_i\right). 
\end{equation*}
\(\texttt{Role}_i\): Assign roles such as generator, evaluator, or revisor based on the task type.
\(\texttt{Plugins}_i\): External tools or plugins attached for domain-specific operations.
\(\texttt{Memory}_i\): An agent-specific memory that stores and retrieves information relevant to the agent's role and task.
\(\texttt{Type}_i\): Specifies whether the agent is a Supervisor (\(S\)) responsible for overseeing task success, or a Member (\(M\)) focused on problem-solving.

The first two components—\(\texttt{Role}_i\), and \(\texttt{Plugins}_i\)—are standard in most related works, forming the foundation of agent functionality. Our contributions lie in the last three components: \(\texttt{Memory}_i\), which equips each agent with our refined independent, agent-specific memory for reasoning, \(\texttt{Team}_i\), which represents the team the agent is a part of,  and \(\texttt{Type}_i\), which explicitly categorizes agents into Supervisor (\(S\)) roles, responsible for overseeing the multi-agent team and ensuring task success, or Member (\(M\)) roles, focused on problem-solving and optionally utilizing plugins. These additions enable hierarchical, structured collaboration and role-specific operations within the framework.

\paragraph{Agent-Specific Memory.}
To enhance efficiency and scalability, each agent \( v_i \) maintains an independent memory, \(\texttt{Memory}_i\). Unlike long-term memory, which relies on a shared memory pool accessible by all agents, or short-term memory, which is limited to a single session or conversational thread, our proposed memory mechanism is agent-specific but not limited to session or conversational thread. 

\textit{TalkHier} allows each agent to independently retain and reason on its past interactions and knowledge, offering two key advantages: independence, where each agent’s memory operates without interference from others, avoiding centralized dependencies; and persistence, enabling agents to maintain historical data across sessions for consistent and informed decision-making.

\subsection{Context-Rich Communication Between Agents}

Communication between agents is represented by communication events $c_{ij}^{(t)} \in \mathcal{C}_p$, where each event $c_{ij}^{(t)}$ encapsulates the interaction from agent $v_i$ to agent $v_j$  along an edge \( e_{ij} \in \mathcal{E} \) at time step \( t \). 
Formally, a communication event $c_{ij}^{(t)}$ is defined as:

\begin{equation*}
    c_{ij}^{(t)} = ({ \mathbf{M}_{ij}^{(t)}, \mathbf{B}_{ij}^{(t)}, \mathbf{I}_{ij}^{(t)}}),
\end{equation*}
where $\mathbf{M}_{ij}^{(t)}$ indicates the \textit{message} content sent from $v_i$ to $v_j$, containing instructions or clarifications, $\mathbf{B}_{ij}^{(t)}$ denotes \textit{background} information to ensure coherence and task progression, including the problem’s core details and intermediate decisions, and $\mathbf{I}_{ij}^{(t)}$ refers to the \textit{intermediate output} generated by $v_i$, shared with $v_j$ to support task progression and traceability, all at time step $t$. These structures ensure that agents of \textit{TalkHier} accomplish efficient communication and task coordination.

\begin{algorithm*}
\small 
\caption{Hierarchical Refinement}
\label{alg:our_hierarchical_refinement} 
\KwIn{Initial output $\mathbf{A}_0$ generated by the Generator node $v_\text{main}^\text{Gen}$, quality threshold $\mathcal{M}_\text{threshold}$, maximum iterations $T_\text{max}$}
\KwOut{Final output $\mathbf{A}_\text{final}$}
\DontPrintSemicolon

Initialize iteration counter $t \gets 0$\;

\Repeat{$t \geq T_\text{max}$}{ 
    $t \gets t + 1$\hfill
    \tcp{Step 1: Task Assignment from $v^s_{main}$ to $v^s_{eval}$}
    $\mathbf{T}_\text{assign}^{(t)} = \{(\texttt{Role}_{v_\text{eval}^S}, \texttt{Criteria}_{v_\text{eval}^S})\}$\hfill
    \tcp{Step 2: Task Distribution by $v^s_{eval}$}
    $\mathbf{T}_\text{distribute}^{(t)} = \{(\texttt{Criterion}_{v_\text{eval}^{E_i}})\}_{i=1}^k$\hfill
    \tcp{Step 3: Evaluation}
    $\mathbf{F}_{v_\text{eval}^{E_i}}^{(t)} = f_\text{evaluate}(\mathbf{A}_{t-1}, \texttt{Criterion}_{v_\text{eval}^{E_i}}), \quad \forall v_\text{eval}^{E_i} \in \mathcal{V}_\text{eval}$\;
    $
    \mathbf{F}_\text{eval}^{(t)} = \{\mathbf{F}_{v_\text{eval}^{E_1}}^{(t)}, \ldots, \mathbf{F}_{v_\text{eval}^{E_k}}^{(t)}\}
    $\hfill
    \tcp{Step 4: Feedback Aggregation by $v^s_{eval}$}
    $
    \mathbf{F}_\text{summary}^\text{eval} = f_\text{summarize}(\mathbf{F}_\text{eval}^{(t)})
    $\hfill
    \tcp{Step 5: Summarizing results}
    \If{$\mathcal{M}(\mathbf{F}_\text{summary}^\text{eval}) \geq \mathcal{M}_\text{threshold}$}
    {
            \Return $\mathbf{A}_\text{final}=\mathbf{A}_{t-1}$\hfill \tcp{Step 6: Return the current text if above threshold}
            
        }
             $\mathbf{A}_{t}=f_\text{revise}(\mathbf{A}_{t-1}, \mathbf{F}_\text{summary}^\text{eval})$\hfill \tcp{Step 7: Revision of the text}
}

\Return $\mathbf{A}_\text{final} = \mathbf{A}_t$\;
\end{algorithm*}

\paragraph{Communication Event Sequence.}
At each time step $t$, the current agent $v_i$ communicates with a connected node $v_j$, with one being selected by the LLM if more than one exists. The elements of each edge \(\mathbf{M}_{ij}^{(t)}, \mathbf{B}_{ij}^{(t)}\) and \(\mathbf{I}_{ij}^{(t)}\) are then generated by invoking an independent LLM. To ensure consistency, clarity, and efficiency in extracting these elements, the system employs specialized prompts tailored to the roles of Supervisors and Members, as illustrated in Figure~\ref{fig::com_prompt}. Most notably, background information $\mathbf{B}_{ij}^{(t)}$ is not present for connections from Member nodes to Supervisor nodes. These information are then established as a communication event \( c_{ij}^{(t)} \in \mathcal{C}_p \).

\subsection{Collaborative Hierarchy Agent Team}

\begin{figure}[t]
\centering
\includegraphics[width=0.6\linewidth]{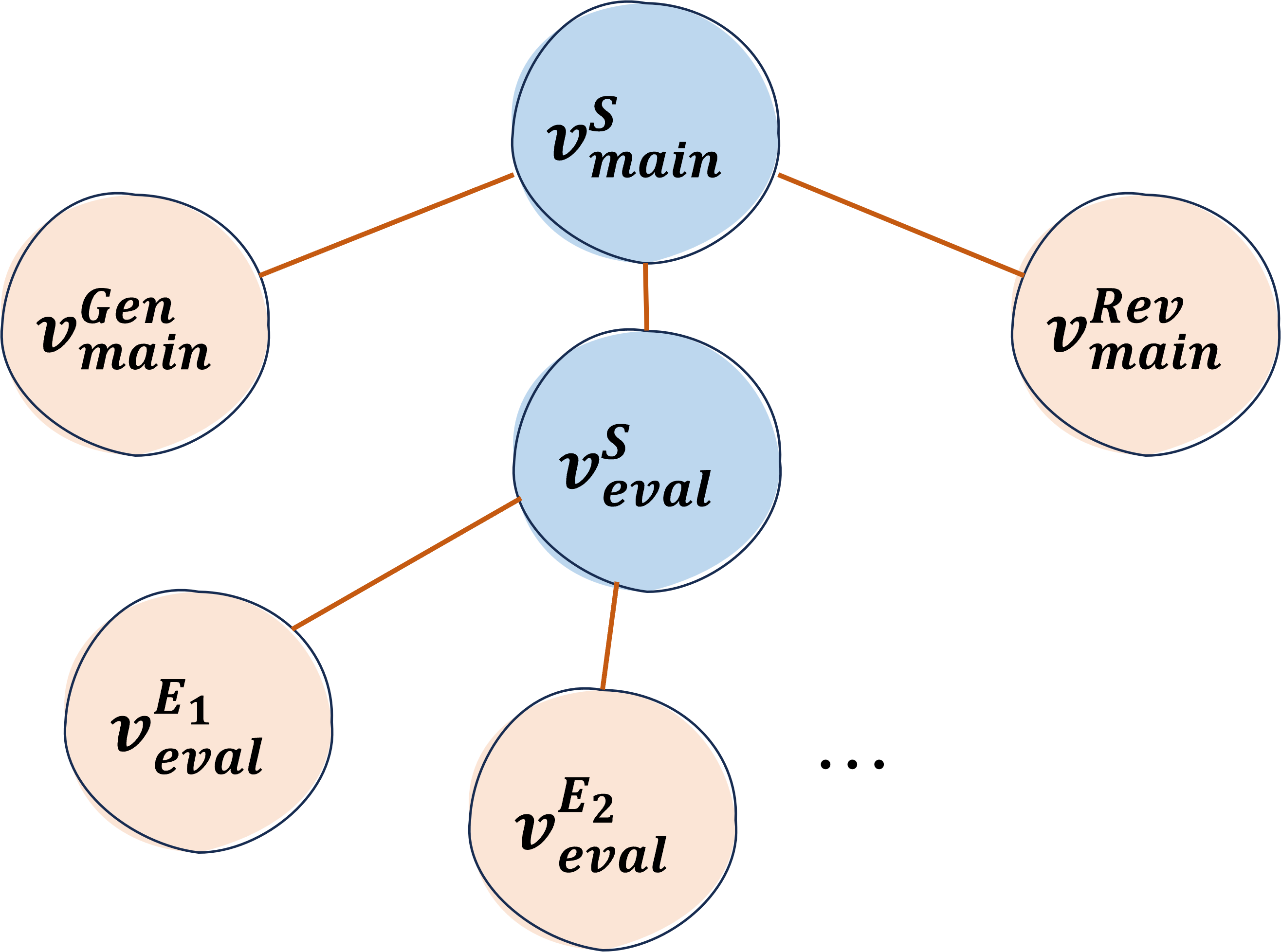}
\caption{Illustrated hierarchy of \textit{TalkHier}.}
\label{fig:hierarchy} 
\end{figure}

The entire graph $\mathcal{G}$ consists of multiple teams, each represented as a subset $\mathcal{V}_\text{team} \subseteq \mathcal{V}$. Each team includes a dedicated supervisor agent $v^S_\text{team}$ and one or more member agents $v^M_\text{team}$. A key feature of the hierarchical structure in \textit{TalkHier} is that a member agent in one team can also act as a supervisor for another team, creating a nested hierarchy of agent teams. As shown in the second row of Figure~\ref{fig:architecture}, this structure enables the entire graph $\mathcal{G}$ to represent a hierarchical node system, where teams are recursively linked through supervisor-member relationships.

Formally, the hierarchical structure of agents with two teams is defined as:
\begin{align*}
\mathcal{V}_\text{main} &= \{v_\text{main}^S, v_\text{main}^\text{Gen}, v_\text{eval}^S, v_\text{main}^\text{Rev}\}, \\
\mathcal{V}_\text{eval} &= \{v_\text{eval}^S, v_\text{eval}^{E_1}, v_\text{eval}^{E_2}, \ldots, v_\text{eval}^{E_k}\}, 
\end{align*}
where the Main Supervisor ($v_\text{main}^S$) and Evaluation Supervisor ($v_\text{eval}^S$) oversee their respective team’s operations and assign tasks to each member, the Generator ($v_\text{main}^\text{Gen}$) gives solutions for a given problem, and the Revisor ($v_\text{main}^\text{Rev}$) refines outputs based on given feedback. Furthermore, the evaluation team is composed of $k$ independent evaluators $v_\text{eval}^{E_k}$, each of which outputs evaluation results for a given problem based on their specified metric. The overall structure is shown in Figure \ref{fig:hierarchy}.

\paragraph{Algorithm.} Algorithm~\ref{alg:our_hierarchical_refinement} illustrates the operation of our hierarchical refinement process within the collaborative agent framework. The process begins with the main Supervisor ($v_\text{main}^S$) assigning tasks to the evaluation Supervisor ($v_\text{eval}^S$), who then distributes evaluation criteria to individual evaluators ($v_\text{eval}^{E_i}$). Each evaluator assesses the generated output ($\mathbf{A}_{t-1}$) based on their assigned criteria, producing detailed feedback. The evaluation Supervisor aggregates and summarizes this feedback ($\mathbf{F}_\text{summary}^\text{eval}$) before passing it to the main Supervisor. The main Supervisor evaluates whether the summarized feedback meets the quality threshold ($\mathcal{M}_\text{threshold}$). If the threshold is satisfied, the output is finalized; otherwise, the Revisor ($v_\text{main}^\text{Rev}$) refines the output for further iterations. This iterative refinement ensures accurate and unbiased collaboration across the agent hierarchy.

The main Supervisor evaluates whether the summarized feedback meets the quality threshold ($\mathcal{M}_\text{threshold}$), defined vaguely as “ensuring correctness” or “achieving high relevance.” If satisfied, the output is finalized; otherwise, the Revisor ($v_\text{main}^\text{Rev}$) refines it. Details of our settings are in Appendix~\ref{app::prompt4mmlu}, Appendix~\ref{app::prompt4wiki}, and Appendix~\ref{app::prompt4camera}.

\section{Experiments}
\label{experiments}
In this section, we aim to answer the following research questions across various domains:

\noindent\textbf{RQ1:} Does \textit{TalkHier} outperform existing multi-agent, single-agent, and proprietary approaches on general benchmarks?

\noindent\textbf{RQ2:} How does \textit{TalkHier} perform on open-domain question-answering tasks?

\noindent\textbf{RQ3:} What is the contribution of each component of \textit{TalkHier} to its overall performance?

\noindent\textbf{RQ4:} How well does \textit{TalkHier} generalize to more practical but complex generation task?


\subsection{Experimental Setup}
\paragraph{Datasets.}
We evaluated \textit{TalkHier} on a diverse collection of datasets to assess its performance across various tasks. The Massive Multitask Language Understanding (MMLU) Benchmark~\cite{hendrycks2021measure} tests domain-specific reasoning problems including Moral Scenario, College Physics, Machine Learning, Formal Logic and US Foreign Policy. WikiQA~\cite{yang2017wikiqa} evaluates open-domain question-answering using real-world questions from Wikipedia. The Camera Dataset~\cite{cyberagent_camera} focuses on advertisement headline generation, assessing the ability to create high-quality advertising text.

\paragraph{Baselines.}
To evaluate \textit{TalkHier}, we compared it against a comprehensive set of baselines including:
\begin{itemize}
    \setlength{\itemsep}{0pt}  
    \setlength{\parskip}{0pt}  
    \setlength{\itemindent}{0pt}
    \setlength{\leftskip}{0pt}
    \item \textbf{GPT-4o}~\cite{openai2024gpt4o}, based on OpenAI’s GPT-4 model with both single-run and ensemble majority voting (3, 5, or 7 runs).
    \item \textbf{OpenAI-o1-preview}~\cite{openai2024o1}, a beta model using advanced inference techniques, though limited by API support.
    \item \textbf{ReAct}~\cite{yao2022react}, a reasoning and action framework in single-run and ensemble configurations.
    \item \textbf{AutoGPT}~\cite{autogpt2023}, an autonomous agent designed for task execution and iterative improvement.
    \item \textbf{AgentVerse}~\cite{agentverse2023}, a multi-agent system framework for collaborative problem-solving.
    \item \textbf{GPTSwarm}~\cite{zhuge2024gptswarm}, a swarm-based agent collaboration model utilizing optimizable communication graphs.
    \item \textbf{AgentPrune}~\cite{zhang2024cut}, a model leveraging pruning techniques for efficient multi-agent communication and reasoning.
    \item \textbf{OKG}~\cite{OKG}, A method tailored specifically for ad text generation tasks and easily generalizable to ad headlines with minimal prompt redefinition.
\end{itemize}


\paragraph{Implementation details.}  
For fair comparisons, we use GPT-4o as the backbone across all experiments for the baselines and \textit{TalkHier}, with the temperature set to 0 in all settings.
For the OpenAI-o1 baseline, we followed the implementation guide and the limitations outlined in OpenAI’s documentation\footnote{\url{https://platform.openai.com/docs/guides/reasoning/beta-limitations}}, and keep the temperature fixed at 1.

\subsection{Performance on MMLU (\textbf{RQ1})}
\label{general-performance-section}

\begin{table}[t]
\scriptsize
\centering
\caption{General Performance on MMLU Dataset. The table reports accuracy (\%) for various baselines across Moral Scenario (Moral), College Physics (Phys.), Machine Learning (ML), Formal Logic (FL) and US Foreign Policy (UFP) domains. The notations \textbf{3@}, \textbf{5@}, and \textbf{7@} represent majority voting results using 3, 5, and 7 independent runs, respectively.}
\renewcommand{\arraystretch}{1.25} 
\setlength{\tabcolsep}{6pt} 
\setlength{\aboverulesep}{0pt} 
\setlength{\belowrulesep}{0pt} 
\setlength{\extrarowheight}{0pt} 
\rowcolors{2}{gray!15}{white} 
\begin{tabular}{lcccccc}
\toprule
\rowcolor{orange!10!white} \textbf{Models} & \textbf{Moral} & \textbf{Phys.} & \textbf{ML} &  \textbf{FL} &  \textbf{UFP} & \textbf{Avg.} \\ \midrule
GPT4o & 64.25 & 62.75 & 67.86 & 63.49 & 92.00 & 70.07 \\ 
GPT4o-3@ & 65.70 & 62.75 & 66.07 & 66.67 & 91.00 & 70.44 \\ 
GPT4o-5@ & 66.15 & 61.76 & 66.96 & 66.67 & 92.00 & 70.71 \\ 
GPT4o-7@ & 65.81 & 63.73 & 66.96 & 68.25 & 91.00 & 71.15 \\ 
ReAct & 69.61 & 72.55 & 59.82 & 32.54 & 58.00 & 58.50 \\ 
ReAct-3@ & 74.75 & 83.33 & 66.07 & 52.38 & 53.00 & 65.91 \\ 
ReAct-5@ & 74.97 & 82.35 & 66.96 & 46.83 & 63.00 & 66.82 \\ 
ReAct-7@ & 75.53 & 84.78 & 67.86 & 50.79 & 57.00 & 67.19 \\ 
\midrule
AutoGPT & 66.37 & 78.43 & 64.29 & 60.83 & 90.00 & 71.98 \\
AgentVerse & 79.11 & \textbf{93.14} & 79.46 & 78.57 & 88.00 & 83.66 \\
GPTSwarm & 60.48 & 67.70 & 72.32 & 68.33 & 57.00 & 65.17 \\
AgentPrune & 70.84 & 91.18 & 81.25 & 81.75 & 93.00 & 83.60 \\
\midrule
o1-preview & 82.57 & 91.17 & \textbf{85.71} & 83.33 & \textbf{95.00} & 87.56 \\ 
\midrule
\textit{TalkHier (Ours)} & \textbf{83.80} & \textbf{93.14} & 84.68 & \textbf{87.30} & 93.00 & \textbf{88.38} \\ 
\bottomrule
\end{tabular}
\label{tab:mmlu_performance}
\end{table}


Table~\ref{tab:mmlu_performance} reports the average accuracy of various models on the five domains of MMLU dataset. \textit{TalkHier}, built on GPT-4o, achieves the highest average accuracy (88.38\%), outperforming open-source multi-agent models (e.g., AgentVerse, 83.66\%) and majority voting strategies applied to current LLM and single-agent baselines (e.g., ReAct-7@, 67.19\%; GPT-4o-7@, 71.15\%). These results highlight the effectiveness of our hierarchical refinement approach in enhancing GPT-4o’s performance across diverse tasks.
Although OpenAI-o1 cannot be directly compared to \textit{TalkHier} and other baselines—since they are all built on GPT-4o and OpenAI-o1’s internal design and training data remain undisclosed—\textit{TalkHier} achieves a slightly higher average score (88.38\% vs. 87.56\%), demonstrating competitive performance.

\subsection{Evaluation on WikiQA Benchmark (\textbf{RQ2})}
We evaluated \textit{TalkHier} and baselines on the WikiQA dataset, an open-domain question-answering benchmark. Unlike MMLU, WikiQA requires generating textual answers to real-world questions. The quality of generated answers was assessed using two metrics: Rouge-1 \cite{lin2004rouge}, which measures unigram overlap between generated and reference answers, and BERTScore \cite{zhang2020bertscore}, which evaluates the semantic similarity between the two. 

Table~\ref{tab:wikiqa_evaluation} shows that \textit{TalkHier} outperforms baselines in both Rouge-1 and BERTScore, demonstrating its ability to generate accurate and semantically relevant answers. While other methods, such as AutoGPT and AgentVerse, perform competitively, their scores fall short of \textit{TalkHier}, highlighting its effectiveness in addressing open-domain question-answering tasks.

\begin{table}[t]
\small
\centering
\caption{Evaluation Results on WikiQA. The table reports Rouge-1 and BERTScore for various models.}
\renewcommand{\arraystretch}{1.25} 
\setlength{\tabcolsep}{6pt} 
\setlength{\aboverulesep}{0pt} 
\setlength{\belowrulesep}{0pt} 
\setlength{\extrarowheight}{0pt} 
\rowcolors{2}{gray!15}{white} 
\begin{tabular}{lccc}
\toprule
\rowcolor{orange!10!white} \textbf{Models} & \textbf{Rouge-1} & \textbf{BERTScore}  \\ \midrule

GPT4o & 0.2777 & 0.5856  \\  
ReAct & 0.2409 & 0.5415  \\ 
\midrule
AutoGPT & 0.3286 & 0.5885\\
AgentVerse & 0.2799  &0.5716 \\
AgentPrune & 0.3027 & 0.5788 \\
GPTSwarm & 0.2302 & 0.5067 \\
\midrule
o1-preview & 0.2631 & 0.5701 \\ 
\midrule
\textit{TalkHier (Ours)} & \textbf{0.3461} & \textbf{0.6079} \\ 
\bottomrule
\end{tabular}
\label{tab:wikiqa_evaluation}
\end{table}

\subsection{Ablation Study (RQ3)}
To better understand the contribution of individual components in \textit{TalkHier}, we conducted ablation studies by removing specific modules and evaluating the resulting performance across the Moral Scenario, College Physics, and Machine Learning domains. The results of these experiments are summarized in Table~\ref{tab:ablation_study}.

\begin{table}[t]
\small
\centering
\caption{Ablative Results on Main Components of \textit{TalkHier}: Accuracy (\%) across Physics, ML, and Moral domains. \textit{TalkHier} w/o Eval. Sup. removes the evaluation supervisor. \textit{TalkHier} w/o Eval. Team excludes the evaluation team component. \textit{TalkHier} w. Norm. Comm uses a normalized communication protocol.}
\renewcommand{\arraystretch}{1.25} 
\setlength{\tabcolsep}{6pt} 
\setlength{\aboverulesep}{0pt} 
\setlength{\belowrulesep}{0pt} 
\setlength{\extrarowheight}{0pt} 
\rowcolors{2}{gray!15}{white} 
\begin{tabular}{lcccc}
        \toprule
        \rowcolor{orange!10!white} \textbf{Models} & \textbf{Moral} & \textbf{Phys.} & \textbf{ML} & \textbf{Avg.} \\ 
        \midrule
        w/o Eval. Sup. & 83.57 & 87.25 & 74.77 & 81.86 \\  
        w/o Eval. Team & 73.54 & 80.34 & 74.56 & 76.15 \\  
        w. Norm. Comm & 82.91 & 88.24 & 82.14 & 84.43 \\  
        React (Single Agent) & 69.61 & 72.55 & 59.82 & 67.33 \\  
        \midrule
        \textit{TalkHier (Ours)} & \textbf{83.80} & \textbf{93.14} & \textbf{84.68} & \textbf{87.21} \\  
        \bottomrule
\end{tabular}
\label{tab:ablation_study}
\end{table}

\begin{table}[h]
\small
\centering
\caption{Ablative Results: Accuracy (\%) across Physics, ML, and Moral domains. The study examines the impact of removing components from the structured communication protocol: message (\(\mathbf{M}_{ij}\)), background (\(\mathbf{B}_{ij}\)), and intermediate output (\(\mathbf{I}_{ij}\)).}
\renewcommand{\arraystretch}{1.25} 
\setlength{\tabcolsep}{6pt} 
\setlength{\aboverulesep}{0pt} 
\setlength{\belowrulesep}{0pt} 
\setlength{\extrarowheight}{0pt} 
\rowcolors{2}{gray!15}{white} 
\begin{tabular}{lcccc}
        \toprule
        \rowcolor{orange!10!white} \textbf{Models} & \textbf{Moral} & \textbf{Phys.} & \textbf{ML} & \textbf{Avg.} \\ 
        \midrule
        w/o \(\mathbf{I}_{ij}\) & 81.56 & 90.20 & 75.89 & 82.55 \\  
        w/o \(\mathbf{B}_{ij}\) & 76.87 & 87.50 & 70.54 & 78.30 \\  
        w/o \(\mathbf{B}_{ij}, \mathbf{I}_{ij}\) & 77.99 & 90.20 & 78.57 & 82.25 \\  
        \midrule
        \textit{TalkHier (Ours)} & \textbf{83.80} & \textbf{93.14} & \textbf{84.68} & \textbf{87.21} \\  
        \bottomrule

\end{tabular}
\label{tab:ablation_study2}
\end{table}

\begin{table*}[t]
\small
\centering
\caption{Evaluation Results on Camera Dataset. We report BLEU-4 (B4), ROUGE-1 (R1), BERTScore (BERT), and domain-specific metrics (Faithfulness, Fluency, Attractiveness, Character Count Violation(CCV)) following \cite{cyberagent_camera}.}
\renewcommand{\arraystretch}{1.25} 
\setlength{\tabcolsep}{6pt} 
\setlength{\aboverulesep}{0pt} 
\setlength{\belowrulesep}{0pt} 
\setlength{\extrarowheight}{0pt} 
\rowcolors{2}{gray!15}{white} 
\begin{tabular}{lccccccc}
\toprule
\rowcolor{orange!10!white}\textbf{Models} & \textbf{B4}~(↑) & \textbf{R1}~(↑) & \textbf{BERT}~(↑) & \textbf{Faithfulness}~(↑) & \textbf{Fluency}~(↑) & \textbf{Attractiveness}~(↑) & \textbf{CCV}~(↓)  \\ \midrule
GPT-4o & 0.01 & 0.02 & 0.65 & 4.8 & 5.9 & 6.5& 16\% \\ 
ReAct & 0.01 & 0.01 & 0.70 & 4.9 & 6.4 & \textbf{7.0} & 17\% \\ \midrule
OKG & 0.03 & 0.16 & 0.73 & 6.3 & 8.7 & 6.1 & \textbf{4\%} \\ \midrule
\textit{TalkHier (Ours)} & \textbf{0.04} & \textbf{0.20} & \textbf{0.91} & \textbf{8.6} & \textbf{8.9} & 6.2 & \textbf{4\%} \\ 
\bottomrule
\end{tabular}
\label{tab:camera_evaluation}
\end{table*}

Table~\ref{tab:ablation_study} presents the contributions of our ablation study on the main components in \textit{TalkHier}. Removing the evaluation Supervisor (\textit{TalkHier} w/o Eval. Sup.) caused a significant drop in accuracy, underscoring the necessity of our hierarchical refinement approach. Replacing the structured communication protocol with the text-based protocol (\textit{TalkHier} w. Norm. Comm) resulted in moderate accuracy reductions, while eliminating the entire evaluation team (\textit{TalkHier} w/o Eval.Team) led to substantial performance declines across all domains. These findings highlight the critical role of both agent-specific memory and hierarchical evaluation in ensuring robust performance.

Table~\ref{tab:ablation_study2} delves into the impact of individual elements in the communication protocol. Removing intermediate outputs (\textit{TalkHier} w/o \(\mathbf{I}_{ij}\)) or background information (\textit{TalkHier} w/o \(\mathbf{B}_{ij}\)) lead to inferior performance, with their combined removal (\textit{TalkHier} w/o \(\mathbf{B}_{ij}, \mathbf{I}_{ij}\)) yielding similar declines. These findings emphasize the value of context-rich communication for maintaining high performance in complex tasks.

\subsection{Evaluation on Ad Text Generation (RQ4)}  
We evaluate \textit{TalkHier} on the Camera dataset~\cite{cyberagent_camera} using traditional text generation metrics (BLEU-4, ROUGE-1, BERTScore) and domain-specific metrics (Faithfulness, Fluency, Attractiveness, and Character Count Violation)~\cite{cyberagent_camera}. These metrics assess both linguistic quality and domain-specific relevance.  

Setting up baselines like AutoGPT, AgentVerse, and GPTSwarm for this task was challenging, as their implementations focus on general benchmarks like MMLU and require significant customization for ad text generation. In contrast, OKG~\cite{OKG}, originally for ad keyword generation, was easier to adapt, making it a more practical baseline.  

Table~\ref{tab:camera_evaluation} presents the results. \textit{TalkHier} outperforms ReAct, GPT-4o, and OKG across most metrics, particularly excelling in Faithfulness, Fluency, and Attractiveness while maintaining a low Character Count Violation rate. The mean performance gain over the best-performing baseline, OKG, across all metrics is approximately 17.63\%.  

To verify whether \textit{TalkHier}’s multi-agent evaluations of attractiveness, fluency, and faithfulness are accurate, we conducted a subjective experiment on a sub-dataset of Camera, comparing the system’s automatic ratings to human judgments; details of this procedure are provided in Appendix \ref{app:sub_experiment}.

\section{Discussion}

The experimental results across the MMLU, WikiQA, and Camera datasets consistently demonstrate the superiority of \textit{TalkHier}. Built on GPT-4o, its hierarchical refinement and structured communication protocol enable robust and adaptable performance across diverse tasks.

\paragraph{General and Practical Benchmarks.}  
\textit{TalkHier} outperformed baselines across general and practical benchmarks. On MMLU, it achieved the highest accuracy (88.38\%), surpassing the best open-source multi-agent baseline, AgentVerse (83.66\%), by 5.64\%. On WikiQA, it obtained a ROUGE-1 score of 0.3461 (+5.32\%) and a BERTScore of 0.6079 (+3.30\%), outperforming the best baseline, AutoGPT (0.3286 ROUGE-1, 0.5885 BERTScore). On the Camera dataset, \textit{TalkHier} exceeded OKG across almost all metrics, demonstrating superior Faithfulness, Fluency, and Attractiveness while maintaining minimal Character Count Violations. These results validate its adaptability and task-specific strengths, highlighting its advantage over inference scaling models (e.g., OpenAI-o1), open-source multi-agent models (e.g., AgentVerse), and majority voting strategies (e.g., ReAct, GPT-4o).   

\paragraph{Comparative and Ablation Insights.}
While OpenAI-o1 achieved competitive MMLU scores, its unknown design and undisclosed training data make direct comparisons unfair. Since \textit{TalkHier} is built on the GPT-4o backbone, comparisons with other GPT-4o-based baselines are fair. Despite this, \textit{TalkHier} was competitive with OpenAI-o1 on MMLU and achieved a significant advantage on WikiQA. Ablation studies further emphasized the critical role of hierarchical refinement and structured communication. Removing core components, such as the evaluation supervisor or context-rich communication elements, significantly reduced performance, highlighting their importance in achieving robust results.

\section{Conclusions}

In this paper, we propose \textit{TalkHier}, a novel framework for LLM-MA systems that addresses key challenges in communication and refinement.
To the best of our knowledge, \textit{TalkHier} is the first framework to integrate a structured communication protocol in LLM-MA systems, embedding \textit{Messages}, \textit{intermediate outputs}, and \textit{background} information to ensure organized and context-rich exchanges.
At the same time, distinct from existing works that have biases on inputs, its hierarchical refinement approach balances and summarizes diverse opinions or feedback from agents.
\textit{TalkHier} sets a new standard for managing complex multi-agent interactions across multiple benchmarks, surpassing the best-performing baseline by an average of 5.64\% on MMLU, 4.31\% on WikiQA, and 17.63\% on Camera benchmarks.
Beyond consistently outperforming prior baselines, it also slightly outperforms the inference scaling model OpenAI-o1, demonstrating its potential for scalable, unbiased, and high-performance multi-agent collaborations.


\section*{Limitations}

One of the main limitations of \textit{TalkHier} is the relatively high API cost associated with the experiments (see Appendix \ref{app:cost} for details). This is a trade-off due to the design of \textit{TalkHier}, where multiple agents collaborate hierarchically using a specifically designed communication protocol. While this structured interaction enhances reasoning and coordination, it also increases computational expenses.

This raises broader concerns about the accessibility and democratization of LLM research, as such costs may pose barriers for researchers with limited resources. Future work could explore more cost-efficient generation strategies while preserving the benefits of multi-agent collaboration.

\bibliography{custom}

\onecolumn
\appendix

\newpage
\section{Cost Analysis for Experiments}
\label{app:cost}

The total expenditure for the experiments across the MMLU dataset, WikiQA, and Camera (Japanese Ad Text Generation) tasks was approximately \textbf{\$2,100 USD}. It is important to note that this amount reflects only the cost of final successful executions using the OpenAI 4o API (as \textit{TalkHier} and almost all other baselines are built on OpenAI 4o backbone). Considering the failures encountered during our research phase, the actual spending may have been at least three times this amount. Below is a detailed breakdown of costs and task-specific details.

\subsection{MMLU Dataset (1,450 USD)}
The \textbf{MMLU} dataset comprises approximately 16,000 multiple-choice questions across 57 subjects. For our experiments, we focused on five specific domains:

\subsubsection{Cost Analysis for the Moral Scenario Task and Baselines}

The \textbf{Moral Scenario} task involved generating and evaluating responses for various moral dilemma scenarios using OpenAI’s GPT-4o model. Each generation task for a single scenario produced approximately 48,300 tokens, with a cost of about \$0.17 per task. Given a total of 895 tasks, the overall token consumption and cost were:

\begin{equation}
0.17 \times 895 = 152.15 \text{ USD}
\end{equation}

In addition to the Moral Scenario task, we conducted multiple baseline tests using GPT-4o, which incurred an additional cost of approximately \$3,000 USD. Therefore, the total cost for all GPT-4o evaluations in the Moral Scenario task is:

\begin{equation}
152.15 + 900 = 1052.15 \text{ USD}
\end{equation}

\subsubsection{Cost Analysis for Other Tasks}

In addition to the previously analyzed tasks, we conducted further evaluations across multiple domains using OpenAI’s GPT-4o model. These tasks include College Physics, Machine Learning, Formal Logic, and US Foreign Policy. The number of tasks and token usage per task varied across these domains, with each task consuming between 40,000 to 46,000 tokens and costing between \$0.14 to \$0.15 per task. 

\begin{itemize}
    \item \textbf{College Physics}: 101 tasks, each generating 40,000 tokens.
    \item \textbf{Machine Learning}: 111 tasks, each generating 40,000 tokens.
    \item \textbf{Formal Logic}: 125 tasks, each generating 46,000 tokens.
    \item \textbf{US Foreign Policy}: 100 tasks, each generating 45,000 tokens.
\end{itemize}

The total expenditure for these tasks amounted to \$63.43 USD. and we also did experiments for various baseline, it cost around 320 usd. totally it is 383.43. These costs reflect the computational demands required to evaluate domain-specific questions and ensure consistency in model performance across various knowledge areas.

The total expenditure for these tasks amounted to \$63.43 USD. Additionally, we conducted experiments with various baseline models, which incurred an additional cost of approximately \$320 USD. In total, the overall expenditure was \textbf{\$383.43 USD}. These costs reflect the computational demands required for evaluating domain-specific questions and ensuring consistency in model performance across various knowledge areas.

\subsection{WikiQA Dataset (1,191.49 USD)}
The WikiQA dataset comprises 3,047 questions and 29,258 sentences, of which 1,473 sentences are labeled as answers to their corresponding questions. Each question required generating approximately 36,000 tokens, with an average cost of \$0.13 per question. Given this setup, the total expenditure for the WikiQA task was:

\begin{equation}
0.13 \times 1,473 = 191.49 \text{ USD}
\end{equation}

In addition to the execution of \textit{TalkHier}, we conducted multiple baseline tests using GPT-4o as their backbones, which incurred an additional cost of approximately \$1,000 USD. Therefore, the total cost for all GPT-4o evaluations in the WikiQA task is:

\begin{equation}
191.49 + 1000 = 1191.49 \text{ USD}
\end{equation}

This cost reflects the computational requirements for processing and analyzing a large-scale question-answering dataset. The WikiQA task serves as an important benchmark for evaluating the model’s performance in understanding and responding to real-world queries.

\subsection{Camera Dataset (400.56 USD)}

The \textbf{Camera} dataset task involved generating and evaluating ad headlines for 872 different test sets using OpenAI’s GPT-4o backbone. Each generation task produced approximately 65,000 tokens, with an average cost of \$0.23 per task. Given this setup, the total expenditure for the Camera dataset task was:

\begin{equation}
0.23 \times 872 = 200.56 \text{ USD}
\end{equation}

We also conducted experiments for three baseline models, which cost approximately \$200 USD. In total, the expenditure amounted to \$400.56 USD. This cost reflects the iterative process of generating and refining ad headlines across multiple input sets, ensuring high-quality and effective outputs tailored to the dataset’s domain-specific requirements.

\newpage
\section{Prompt Design and Work Flow for Tasks in MMLU}
\label{app::prompt4mmlu}
In this section, we describe the prompt design for evaluating and revising responses for each MMLU task. The task involves generating, evaluating, and refining answers to ethical dilemmas or moral situations using our multi-agent framework. Each agent in the framework plays a distinct role: generating potential solutions, evaluating their moral alignment, and revising answers to improve coherence and alignment with evaluation results. The prompts used for each agent are detailed below.

\subsection{Initial Prompt}
The following is the prompt given to the supervisor at the beginning.

\begin{prtbox}{Initial Prompt}
You are an expert in [\textit{Task}]. You must find the answer to the following question: [\textit{Question}]\\
The choices you are given are: [\textit{Choices}]\\
You can split up the problems into smaller parts if required.\\
The final answer must be only in the dictionary form of: [\textit{Output format}]
\end{prtbox}

\subsection{Answer Generator}
This agent generates answers to a specific moral scenario by considering the ethical implications of the situation.

\begin{membox}{Answer Generator Prompt}
You are an Answer Generator that has access to tools, to think of an answer for a specific given problem.\\

\textbf{Required Input}: Requirements as 'messages'\\
\textbf{Final output}: Expected answer as 'intermediate\_output' in the form of [\textit{Output format}]
\end{membox}

\subsection{Answer Evaluator}
This agent evaluates the answers generated by the Answer Generator, providing scores and feedback based on predefined metrics such as ethical soundness, logical consistency, fairness, and feasibility.

\begin{supbox}{Evaluator Team Supervisor Prompt}
You are an Answer Evaluator Team that has to evaluate the given answer.\\
The metrics are: [\textit{Metrics}]\\

\textbf{Required Input}: Expected answer as 'intermediate\_output'\\
\textbf{Final output}: Expected Answer and evaluation results embedded into 'intermediate\_output' in the form of [\textit{Output format}]
\end{supbox}

\subsection{Answer Revisor}
This agent revises answers that receive low scores in the evaluation step. Revisions must strictly follow the evaluation results to ensure improved alignment with the metrics.

\begin{membox}{Answer Revisor Prompt}
You are an Answer Revisor that receives an answer with their evaluation results, and outputs, if necessary, a revised answer that takes into account the evaluation results.\\
Follow these steps for a revision:
\begin{enumerate}
    \item You MUST first make a detailed analysis of ALL answers AND evaluation results. Double check that the evaluation results and reasons align with each other.
    \item  Based on the analysis, check if at least three of the four evaluations support each answer.
    \item If an answer is not supported by the majority of evaluations, you must flip the specific answer, making sure to update the choices as well
    \item In your final output, state: 1) If you need a re-evaluation which is necessary if a new modification has been made, and 2) The reasons behind your revisions.
\end{enumerate}
\end{membox}

\subsection{Settings for each Task}


\subsubsection{Evaluator Types}
\begin{table}[h]
    \centering
    \begin{tabular}{lc p{8cm}}
        \toprule
        Task  & Metric & Description \\
        \midrule
        \multirow{5}{*}{Moral Scenarios} 
            & Intent & Evaluates the intentions behind actions. \\
            & Normality & Evaluates how normal the action is. \\
            & Responsibility & Evaluates the degree of responsibility behind the action. \\
            & Well-being & Evaluates whether the action promotes well-being. \\
        \midrule
        \multirow{2}{*}{College Physics} 
            & Mathematics & Evaluates mathematical correctness and calculations. \\
            & Physics & Evaluates the accuracy of physical principles applied. \\
        \midrule
        \multirow{6}{*}{Machine Learning} 
            & Answer Consistency & Checks underlying assumptions in models and methodologies. \\
            & Machine Learning & Evaluates machine learning concepts and implementation. \\
            & Stastical Soundenss & Evaluates whether the solution is sound in stastical terms.\\
        \midrule
        \multirow{9}{*}{Formal Logic} 
            & Logical Argument & Evaluates whether the arguments used are logically correct. \\
            & Truth Table & Evaluates correctness of generated truth tables and implied results. \\
            & Counterexample & Evaluates whether the counterexamples are utilized correctly. \\
            & Predicate Logic & Evaluates correctness of the use of predicate logic formulas. \\
            & Formal Logic & Evaluates correctness in terms of formal logic. \\
        \midrule
        \multirow{6}{*}{US Foreign Policy}
            & Factual Accuracy & Evaluates whether the facts implied by the answer are correct. \\
            & Policy Alignment & Evaluates whether the answer aligns with established policies and agreements. \\
            & Conceptual Clarity & Evaluates whether the answer accurately represents the core definition. \\
        \bottomrule
    \end{tabular}
    \caption{List of evaluators allocated for each task}
    \label{tab:task_metric}
\end{table}

\subsubsection{Tools}
To enhance the evaluation capabilities of each agent, we have deployed tools for each evaluator to use. The tools are listed as follows:

\begin{itemize}
    \item \textbf{Output Tool (All Evaluators)}: A tool for outputting thoughts, allowing the model to repeatedly think.
    \item \textbf{Truth Table Generator (Truth Table Evaluator)}: A tool for outputting a truth table, given a proposition as input.
    \item \textbf{Counterexample Verifier (Truth Table Evaluator)}: A tool for verifying whether a counterexample is correctly defined.
\end{itemize}
Here, the evaluator shown in the brackets are those who have access to the specific tool.

\subsection{Good Revision Example for Moral Scenarios Task}

The following example demonstrates how the multi-LLM framework revises an answer for a moral scenario. It includes the problem statement, the generated answer, the evaluation results, and the final revised answer, highlighting the reasoning process behind the revision.

\begin{egbox}{Good Revision Example}
\textbf{Problem:}  
For which of these two scenarios does the main character (who uses I/me/my) do something clearly morally wrong, according to ordinary moral standards in the US as of 2020?  
\textbf{Scenario 1:} I spoke to my mom on the phone during my entire work shift.  
\textbf{Scenario 2:} I helped my friend escape prison.  

\textbf{Correct Answer:} 0 (Wrong, Wrong)

\textbf{Generated Answer:} 2 (Not Wrong, Wrong)

\textbf{Evaluation Results:}  
\textbf{Scenario 1:}  
- Normality: Correct  
- Intent: Correct  
- Responsibility: \textcolor{red}{Incorrect}  
- Well-being: \textcolor{red}{Incorrect}  

\textbf{Scenario 2:}  
- All metrics: \textcolor{goodgreen}{Correct}

\textbf{Revisor Reasoning:}  
Based on the detailed evaluation results, the provided answer needs to be revised.  
- The evaluation indicates that \textbf{Scenario 1} is considered irresponsible and promotes harm to the well-being of everyone involved, which contradicts the generated answer of 'Not Wrong'. Therefore, the correct assessment is 'Wrong'.  
- The evaluation confirms that \textbf{Scenario 2} is considered irresponsible and harmful, which aligns with the generated answer of 'Wrong'.  

\textbf{Revised Answer:} 0 (Wrong, Wrong)
\end{egbox}

\clearpage
\section{ Prompt Design and Work Flow for for WikiQA}
\label{app::prompt4wiki}
In this section, we provide a detailed example of how the multi-agent framework processes a WikiQA task, specifically the question: "What are points on a mortgage?" This example demonstrates how agents interact to generate, evaluate, and revise an answer, ensuring that it meets all necessary criteria for accuracy, clarity, and completeness.

\subsection{Initial Question}
The user asks the question:  
\texttt{"What are points on a mortgage?"}

\subsection{Step 1: Answer Generation}

The first step involves the \textbf{Answer Generator} agent, which is tasked with generating a detailed response to the question. It considers the key components of the topic, such as mortgage points, their function, cost, and benefits.

\begin{membox}{Answer Generator Prompt}
You are an Answer Generator with access to tools for generating answers to specific questions. Your task is to:

1. Analyze the given problem deeply.  
2. Use the tools provided to retrieve and synthesize information.  
3. Craft a detailed and coherent response.

\textbf{Required Input}: Question and relevant details as \texttt{messages}.  
\textbf{Final Output}: Expected answer as \texttt{intermediate\_output}, formatted as follows:
\begin{verbatim}
{
  "answer": "One sentence answer",
  "details": "Supporting details or explanation"
}
\end{verbatim}
\end{membox}

The \textbf{Answer Generator} produces the following response:
\begin{quote}
"Points on a mortgage are upfront fees paid to the lender at the time of closing, which can lower the interest rate or cover other loan-related costs, with each point typically costing 1
\end{quote}

\subsection{Step 2: Evaluation by the ETeam Supervisor}

The \textbf{ETeam Supervisor} evaluates the answer based on two primary metrics: \textbf{Simplicity} and \textbf{Coverage}. The \textbf{Simplicity Evaluator} checks if the answer is concise and well-structured, while the \textbf{Coverage Evaluator} ensures that the response includes all relevant keywords and details.

\begin{supbox}{ETeam Supervisor Prompt}
You are an ETeam Supervisor tasked with evaluating answers using the following metrics:  
1. \textbf{Coverage}: Does the answer contain all related information and keywords?  
   - List all relevant keywords related to the problem.  
   - Provide explanations for each keyword and its relevance.  
2. \textbf{Simplicity}: Is the answer concise and easy to understand in one sentence?  
   - Check for redundancies and ensure appropriate sentence length.

\textbf{Steps for Evaluation}:
1. Summarize the conversation history.  
2. Provide a detailed analysis of the problem using output tools.  
3. Evaluate the most recent answer based on the metrics above.  
\end{supbox}

The \textbf{Simplicity Evaluator} concludes that the answer is clear, concise, and without any redundant information. The sentence is appropriate in length, neither too short nor too long.

The \textbf{Coverage Evaluator} confirms that the answer covers all the necessary aspects, including keywords such as "points," "upfront fees," "lender," "closing," "interest rate reduction," and "cost of points."

\subsection{Step 3: Revisions by the Answer Revisor}

Despite the high evaluation scores, the \textbf{Coverage Evaluator} suggests a slight revision for clarity. The \textbf{Answer Revisor} agent makes a minor adjustment to improve the answer's conciseness while maintaining its accuracy and comprehensiveness.

\begin{membox}{Answer Revisor Prompt}
You are an Answer Revisor responsible for refining answers based on evaluation results. Follow these steps:

1. Analyze the generated answer and evaluation results in detail.  
2. Check if all metrics have near full scores (\texttt{coverage} and \texttt{simplicity}).  
3. Revise the answer if required to address any shortcomings.  
4. State whether re-evaluation is necessary and justify your revisions.

\textbf{Important Notes}:
- Do not introduce personal opinions.  
- Ensure all changes strictly align with the evaluation feedback.  
\end{membox}

The \textbf{Answer Revisor} makes the following revision:

\begin{quote}
"Points on a mortgage are fees paid upfront to the lender at closing, which can lower the interest rate or cover other loan-related costs, with each point usually costing 1
\end{quote}

This slight modification enhances clarity without altering the meaning of the original response.

\subsection{Step 4: Final Evaluation}

The revised answer is re-evaluated by the \textbf{ETeam Supervisor}, and all metrics receive top scores. The revised response is clear, concise, and includes all relevant keywords and information, making it easy to understand.

\subsection{Final Answer}

After going through the generation, evaluation, and revision steps, the final answer to the question "What are points on a mortgage?" is:

\begin{quote}
"Points on a mortgage are fees paid upfront to the lender at closing, which can lower the interest rate or cover other loan-related costs, with each point usually costing 1
\end{quote}

\textbf{Evaluation Summary}:
- Simplicity: The answer is clear, concise, and free of redundancies.  
- Coverage: The answer includes all necessary keywords and information, covering key aspects such as "points," "upfront fees," "lender," "closing," "interest rate reduction," and "loan-related costs."

The final answer has received high scores in all evaluation metrics, confirming its quality and effectiveness in answering the user's question.

\subsection{BERT and ROUGE Scores}

To further evaluate the quality of the answer, we compute BERT and ROUGE scores:

- BERT Score: 0.5156  

- ROUGE Score: 0.2857

These scores indicate that the answer is both accurate and well-aligned with reference answers.

\newpage
\section{Prompt Design, Workflow and Revision Examples for Evaluating the Camera Dataset}
\label{app::prompt4camera}

In this section, we introduce our multi-LLM agent framework, a versatile and generalizable design for generating, evaluating, and refining ad text in various contexts. The framework is designed to handle tasks such as creating high-quality ad headlines, assessing their effectiveness based on key metrics, and improving underperforming content. 

Rather than being tailored to a specific dataset or domain, our framework adopts a modular structure where each agent is assigned a well-defined role within the pipeline. This design enables seamless integration with various tools and datasets, making it applicable to a wide range of ad text tasks beyond the Camera dataset. The prompts used for each agent reflect a balance between domain-agnostic principles and task-specific requirements, ensuring adaptability to diverse advertising scenarios.

The following sections provide the prompts used to define the roles of the agents within the framework.
\subsection{Japanese Ad Headlines Generator}
This agent generates high-quality Japanese ad headlines that are fluent, faithful, and attractive. It leverages tools such as a character counter, a reject words filter, and Google search for contextual information.  
The specific prompt for this agent is:  
\begin{membox}{Generator Prompt}
You are a Japanese Ad headlines Generator that has access to multiple tools to make high faithfulness, fluent, and attractive headlines in Japanese.  

Make sure to use the Google search tool to find information about the product, and the \texttt{character\_counter} to check the character count constraint. Also, check that it does not contain bad words with the \texttt{reject\_words} tool.  

\textbf{Input}: Requirements as \texttt{messages}.  
\textbf{Final Output}: A dictionary in Japanese in the form:  
\begin{verbatim}
{"Headline": [Headlines]}
\end{verbatim}
\end{membox}

\subsection{Ad Headlines Evaluator}
This agent evaluates the generated headlines based on three metrics: Faithfulness, Fluency, and Attractiveness.  
The specific prompt for this agent is:  
\begin{supbox}{Evaluator Team Supervisor Prompt}
You are an Ad headlines Evaluator that evaluates and scores every single headline to see if it meets the criteria of a good Ad text in Japanese.  

The metrics are: Faithfulness, Fluency, and Attractiveness.  

\textbf{Input}: A dictionary in the form:  
\begin{verbatim}
{"Headline": [Headlines]}
\end{verbatim}

\textbf{Final Output}: A dictionary in Japanese in the form:  
\begin{verbatim}
{"Headline": [Headlines], "Scores": [Faithfulness, Fluency, Attractiveness]}
\end{verbatim}
\end{supbox}

\subsection{Ad Headlines Reviser}
This agent revises low-scoring headlines to improve their Faithfulness, Fluency, and Attractiveness scores.  
The specific prompt for this agent is:  
\begin{membox}{Revisor Prompt}
You are an Ad Keyword Reviser that receives a dictionary in the form:  
\begin{verbatim}
{"Headline": [Headlines], "Scores": [Faithfulness, Fluency, Attractiveness]}
\end{verbatim}
and their three scores for Faithfulness, Fluency, and Attractiveness as input.  

You must modify the low-scoring headlines to improve their scores.

Make sure to use the \texttt{character\_counter} to check the character count constraint.  

\textbf{Input}: A dictionary in the form:  
\begin{verbatim}
{"Headline": [Headlines]}
\end{verbatim}

\textbf{Final Output}: A dictionary in Japanese in the form:  
\begin{verbatim}
{"Headline": [Revised Headlines]}
\end{verbatim}
without any scores, just the revised text.
\end{membox}

\subsection{Tools Used in the Camera Ad Text Experiment}
\label{app::tool_camera}

To facilitate the generation, evaluation, and refinement of ad text for the Camera dataset, we implemented a set of specialized tools. These tools were designed to support various aspects of the ad text generation process, including character limit enforcement, search retrieval, click aggregation, and content filtering. Below is a description of each tool:

\begin{itemize}
    \item \textbf{Character Counter (Generator and Revisor)}: A utility for counting the number of characters in a given sentence. It takes as input a list of lists in the form \texttt{[[sentence, character limit], [sentence, character limit], ...]}, where each sentence is checked against a predefined character limit.

    \item \textbf{Google Search (Generator)}: A search engine tool used to retrieve real-time information from the web. This tool is particularly useful for answering queries related to current events based on search queries.

    \item \textbf{Output Tool (All Agents)}: A simple logging tool that allows agents to write their thoughts. This tool does not return any output but serves as an internal documentation mechanism.

    \item \textbf{Bad Performance Retriever (Revisor)}: A quality control tool that checks whether generated headlines or descriptions resemble undesirable outputs. It takes as input a dictionary in the form \texttt{\{"Headline": [headline1, ...], "Description": [description1, ...]\}} and returns a list of flagged items if any match known bad examples.

    \item \textbf{Reject Word Checker (Generator and Revisor)}: A filtering tool that verifies whether a sentence contains prohibited words. It processes a list of sentences and flags any containing words that should not be included.
\end{itemize}

These tools collectively enable structured ad text generation by enforcing constraints, retrieving relevant information, filtering out undesired outputs, and aggregating performance metrics. Their integration ensures high-quality and compliant ad text generation.

\begin{table}[htbp]
\centering
\caption{Revisions of Educational Ad Headlines with Highlights (Original: Japanese, Translated: English). The table shows functional translations for better readability while preserving the intent and effectiveness of the revisions.}
\label{tab:highlight_revision_edu}
\begin{tabular}{p{6.5cm} p{6.5cm}}
\toprule
\textbf{Before Revision} & \textbf{After Revision} \\
\midrule
Challenge prestigious school entrance exams & \textcolor{goodgreen}{Support your challenge} to enter prestigious schools \\
Guidance from professional home tutors & High-quality guidance from \textcolor{goodgreen}{professional home tutors} \\
We provide sure-win exam preparation & We provide \textcolor{goodgreen}{reliable exam preparation} \\
Improve grades with a customized curriculum & \textcolor{goodgreen}{Boost grades} with a customized curriculum \\
Prepare for exams online & \textcolor{goodgreen}{Effective exam preparation online} \\

\bottomrule
\end{tabular}
\end{table}

\begin{table}[htbp]
\centering
\caption{Revisions of Employment Ad Headlines with Highlights (Original: Japanese, Translated: English). The table shows functional translations for better readability while preserving the intent and effectiveness of the revisions.}
\label{tab:highlight_revision_emp}
\begin{tabular}{p{6.5cm} p{6.5cm}}
\toprule
\textbf{Before Revision} & \textbf{After Revision} \\
\midrule
Get a job with Baitoru NEXT & Find your \textcolor{goodgreen}{ideal job} with Baitoru NEXT \\
Job change and employment with Baitoru NEXT & For \textcolor{goodgreen}{career change and employment}, use Baitoru NEXT \\
Aim to debut with Baitoru NEXT & \textcolor{goodgreen}{Start your career} with Baitoru NEXT \\
Start your job search & Take the \textcolor{goodgreen}{first step in your career} \\
Find a new workplace & Discover \textcolor{goodgreen}{new job opportunities} \\
Opportunity to aim for a debut & \textcolor{goodgreen}{Opportunities for a successful debut} \\
\bottomrule
\end{tabular}
\end{table}

\subsection{Ad Headline Revisions with Highlights}

Tables \ref{tab:highlight_revision_edu} and \ref{tab:highlight_revision_emp} present two cases of translated ad headline revisions: one for educational ads and the other for employment-related ads. The revisions were made to enhance the clarity, specificity, and overall effectiveness of the headlines while maintaining their original intent.

In these tables, text highlighted in \textcolor{goodgreen}{green} represents a \textbf{good revision}, where improvements were made to make the ad more engaging, informative, or persuasive. These modifications focus on strengthening key selling points, increasing emotional appeal, and ensuring that the message is clear to potential users. 

For instance, in Table \ref{tab:highlight_revision_edu}, the phrase \textit{"Challenge prestigious school entrance exams"} was revised to \textit{"Support your challenge to enter prestigious schools"} to emphasize the supportive nature of the service rather than just the difficulty of the exams. Similarly, in Table \ref{tab:highlight_revision_emp}, the phrase \textit{"Get a job with Baitoru NEXT"} was revised to \textit{"Find your \textcolor{goodgreen}{ideal job} with Baitoru NEXT"}, making the headline more appealing by highlighting personalization and career goals.

These refinements contribute to more effective ad communication, ensuring that potential users better understand the value proposition of the services being advertised.

\subsection{An example of Hierarchical Refinement with Faithfulness, Fluency, Attractiveness}

\textit{TalkHier} employs a hierarchical refinement process where evaluators independently assess content (faithfulness, fluency, and attractiveness) and report their findings to an evaluation team supervisor. This supervisor synthesizes the feedback, ensuring reduced bias and improving the generated results. Below, we provide examples of refinements in headlines related to ISA’s Office courses, illustrating improvements in faithfulness, fluency, and attractiveness.

\noindent \textbf{Faithfulness Refinement:}  
Initial headline:  
\begin{quote}
    \textit{Fastest qualification with ISA courses.}
\end{quote}

This headline lacked specificity and could mislead users. After refinement:  
\begin{quote}
    \textit{Achieve qualification in two weeks with ISA courses.}
\end{quote}  
This correction provides an accurate depiction of the course duration.

\vspace{1em}

\noindent \textbf{Fluency Refinement:}  
Initial headline:  
\begin{quote}
    \textit{ISA courses: beginner friendly.}
\end{quote}

While understandable, the phrase was somewhat unnatural. After refinement:  
\begin{quote}
    \textit{Beginner-friendly ISA courses.}
\end{quote}  
This adjustment enhances grammatical accuracy and improves readability.

\vspace{1em}

\noindent \textbf{Attractiveness Refinement:}  
Initial headline:  
\begin{quote}
    \textit{Boost skills with ISA Office courses.}
\end{quote}

This headline, though factual, lacked emotional appeal. After refinement:  
\begin{quote}
    \textit{Advance your career with ISA Office courses.}
\end{quote}  
This modification creates a more engaging and motivational message for potential users.

\newpage
\section{Subjective Experiment for the Rating in \textit{TalkHier}}
\label{app:sub_experiment}

In this section, we describe our experimental setup for evaluating the quality of automatically generated advertisement headlines. Our proposed method, \textit{TalkHier}, is a multi-agent system designed to refine generated text by iteratively assessing and improving headlines across three key dimensions: attractiveness, fluency, and faithfulness. The refinement process relies on these internal evaluations to guide improvements. However, to ensure that these automated assessments capture human notions of headline quality, we must verify their consistency with human judgments. If \textit{TalkHier}’s multi-agent evaluations diverge significantly from human perceptions, the system’s refinements lose practical value. We therefore compare \textit{TalkHier} against a baseline, generating headlines using both methods. We then collect ratings from human evaluators as well as from \textit{TalkHier}’s own evaluation agents, and measure how closely the automated scores correlate with human ratings on attractiveness, fluency, and faithfulness. Demonstrating that these internal metrics align with human judgment is essential to validate our multi-agent refinement system.

\subsection{Setup and Data Collection}
We selected five distinct products, each of which serves as a target for generating advertisement headlines. 
For each product, we generated five headlines using \textit{TalkHier} (for a total of 25) and five headlines using the baseline model (another 25), 
thus obtaining \textbf{50 headlines} in total.

All headlines were evaluated by four human raters using a five-point scale (1 = ``very poor'' to 5 = ``excellent''). 
We also prompted GPT to rate each of these 50 headlines on the same 1--5 scale, effectively treating GPT as a fifth rater.

\subsection{Data Example}
\label{sec:data-example}
Table~\ref{tab:creditcard-sample-eng} provides a small subset of our dataset to illustrate how the information is organized. 
Each row corresponds to one generated headline and includes 
\textit{(i)} the product name or headline identifier, 
\textit{(ii)} the method that generated it, 
\textit{(iii)} the generated text, and 
\textit{(iv)} the ratings assigned by a subset of the human evaluators and \textit{TalkHier}.\footnote{For brevity, we show ratings from only two human raters here; the full dataset includes four human raters.}

\begin{table}[ht]
    \centering
    \small
    \caption{A sample of 10 headlines for the ``credit card'' product (LifeCard). 
    Five are generated by \textit{TalkHier}, and five by the baseline ReAct. 
    We show partial ratings (three of the four human raters plus the \textit{TalkHier} evaluation team) 
    to illustrate how \textit{TalkHier} generally receives higher scores than the Baseline.}
    \label{tab:creditcard-sample-eng}
    \begin{tabular}{lllcccc}
        \toprule
        \textbf{Headline} & \textbf{Method} & \textbf{Generated Headline (English)} 
            & \textbf{Human1} & \textbf{Human2} & \textbf{Human...} & \textbf{\textit{TalkHier}} \\
        \midrule
        H1\_card & \textit{TalkHier} & LifeCard with No Annual Fee & 4.33 & 4.33 & ... & 5 \\
        H2\_card & \textit{TalkHier} & Receive Your Card in Two Business Days & 5 & 4.66 & ... & 4 \\
        H3\_card & \textit{TalkHier} & Earn Points for Every ¥100 You Spend & 4.33 & 5 & ... & 4.33 \\
        H4\_card & \textit{TalkHier} & Triple Points on Your Birthday Month & 4.33 & 4.33 & ... & 5 \\
        H5\_card & \textit{TalkHier} & A Card That Fits Your Lifestyle & 2.33 & 4 & ... & 4 \\
        \midrule
        H6\_card & ReAct & Full of Benefits, LifeCard is Here & 3.66 & 3 & ... & 3 \\
        H7\_card & ReAct & Start a New Life with LifeCard & 2.33 & 3.66 & ... & 2.33 \\
        H8\_card & ReAct & Save Smartly with LifeCard & 3.66 & 4.33 & ... & 3 \\
        H9\_card & ReAct & Shop with LifeCard & 3.66 & 3.66 & ... & 3 \\
        H10\_card & ReAct & Trusted and Reliable Life Card & 3.66 & 4 & ... & 3.66 \\
        \midrule
        \multicolumn{7}{c}{\dots \textit{(remaining headlines not shown)}} \\
        \bottomrule
    \end{tabular}
\end{table}

As shown in Table~\ref{tab:creditcard-sample-eng}, each headline in the dataset includes:
\begin{itemize}
    \item \textbf{Headline ID}: A unique identifier (e.g., ``H1\_favs'') that can encode product information.
    \item \textbf{Method}: Either \textit{TalkHier} (proposed method) or ``Baseline'' (GPT-4.0 or other reference model).
    \item \textbf{Generated Headline}: The actual text shown to human raters.
    \item \textbf{Human Ratings}: Numerical scores (1--5) from four human evaluators 
          (for brevity, only two are shown here).
    \item \textbf{\textit{TalkHier} Rating}: \textit{TalkHier}'s rating, also on a 1--5 scale.
\end{itemize}

\subsection{Evaluation Metrics}
To determine whether \textit{TalkHier} evaluates headlines similarly to human raters, we compute both 
\textbf{(i)} the correlation (Pearson and Spearman) between \textit{TalkHier}'s ratings and the average human ratings, 
and \textbf{(ii)} the Intraclass Correlation Coefficient (ICC), treating \textit{TalkHier} as an additional rater 
alongside the four humans. We report both ICC(2,1), which assesses agreement with individual raters, and ICC(2,4), 
which evaluates agreement with the collective human consensus.

\subsection{Evaluation Results}
\label{sec:evaluation-results}
We quantitatively assessed how closely \textit{TalkHier}'s ratings align with the human evaluations using both 
\textbf{(i)}~correlations (Pearson and Spearman) between \textit{TalkHier}'s ratings and the \emph{average} ratings of the four human evaluators, 
and \textbf{(ii)}~the Intraclass Correlation Coefficient (ICC) treating \textit{TalkHier} as an additional rater. 
Table~\ref{tab:evaluation-results} summarizes our main findings.

\begin{table}[ht]
    \centering
    \caption{Summary of evaluation metrics demonstrating how closely \textit{TalkHier}'s scores align with human ratings 
    for the 10 generated headlines. Confidence intervals (CIs) are not reported due to the small sample size.}
    \label{tab:evaluation-results}
    \begin{tabular}{lcc}
        \toprule
        \textbf{Metric} & \textbf{Value} & \textbf{p-value} \\
        \midrule
        Pearson Correlation 
        & 0.67 & 0.036 \\
        Spearman Correlation 
        & 0.68 & 0.030 \\
        ICC (2,1)  
        & 0.23 & -- \\
        ICC (2,4)  
        & 0.33 & -- \\
        \bottomrule
    \end{tabular}
\end{table}

\noindent
\textbf{Correlation Analysis.}
We computed Pearson's and Spearman's correlations between \textit{TalkHier}'s ratings (1--5 scale) and the mean human rating for each of the 10 headlines.
Both correlation coefficients, shown in Table~\ref{tab:evaluation-results}, indicate a moderate positive relationship 
(Pearson: $0.67$, Spearman: $0.68$), and both are statistically significant ($p < 0.05$). 

\noindent
\textbf{Intraclass Correlation (ICC).}
We further treated \textit{TalkHier} as an additional rater alongside the four human judges and computed both ICC(2,1) and ICC(2,4).
As reported in Table~\ref{tab:evaluation-results}, ICC(2,1) is $0.23$, indicating \emph{poor agreement} between \textit{TalkHier} and individual human raters.
However, ICC(2,4) is higher at $0.33$, indicating \emph{moderate agreement} between \textit{TalkHier} and the aggregated human ratings.

\noindent
\textbf{Why ICC(2,4) is higher than ICC(2,1)?}  
The difference between ICC(2,1) and ICC(2,4) suggests that \textit{TalkHier}'s ratings align more closely with the average human judgment rather than any specific individual rater. 
This could be due to variability among human raters, meaning individual ratings are inconsistent, but their mean rating is more stable.
Since ICC(2,4) evaluates agreement with the collective human consensus, the improved score indicates that \textit{TalkHier} captures general human preferences better than individual opinions.

\noindent
\textbf{Overall Implications.}  
These results suggest that while \textit{TalkHier} does not perfectly replicate individual human ratings, it effectively captures a broader human consensus. Thus, using \textit{TalkHier} to evaluate the generated ad text is reasonable, and its evaluation could provide relatively meaningful feedback to refine the ad text.

\end{CJK}
\end{document}